\pdfoutput=1

\documentclass[11pt]{article}

\usepackage[final]{acl}
\usepackage{times}
\usepackage{latexsym}
\usepackage{enumitem}
\usepackage{booktabs}
\usepackage{subfigure}  
\usepackage{multirow}

\usepackage[T1]{fontenc}

\usepackage[utf8]{inputenc}

\usepackage{microtype}

\usepackage{inconsolata}
\usepackage{graphicx}
\usepackage{tabularx}
\usepackage{xurl}

\usepackage{amsmath}
\usepackage{times}
\usepackage{latexsym}
\usepackage[T1]{fontenc}

\usepackage[utf8]{inputenc}

\usepackage{microtype}

\usepackage{inconsolata}

\usepackage{graphicx}

\title{The Impact of Reasoning Step Length on Large Language Models}

\author{
 \textbf{Mingyu Jin\textsuperscript{1}*},
 \textbf{Qinkai Yu\textsuperscript{2}*},
 \textbf{Shu Dong\textsuperscript{3}},
 \textbf{Haiyan Zhao\textsuperscript{4}},
\\
 \textbf{Wenyue Hua\textsuperscript{1}},
 \textbf{Yanda Meng\textsuperscript{5}},
 \textbf{Yongfeng Zhang\textsuperscript{1}},
 \textbf{Mengnan Du\textsuperscript{4}},
\\
 \textsuperscript{1}Rutgers University,
 \textsuperscript{2}University of Liverpool,
 \textsuperscript{3}Northwestern University,\\
 \textsuperscript{4}New Jersey Institute of Technology,
 \textsuperscript{5}University of Exeter
\\
 \small{
 \texttt{\{mingyu.jin, yongfeng.zhang, wenyue.hua\}\href{mailto:@rutgers.edu}{@rutgers.edu}, \href{mailto:sgqyu9@liverpool.ac.uk}{sgqyu9@liverpool.ac.uk}
 }}
 \\
 \small\texttt{{\href{mailto:dongshu2024@u.northwestern.edu}{dongshu2024@u.northwestern.edu}, \href{mailto:Y.M.Meng@exeter.ac.uk}{Y.M.Meng@exeter.ac.uk}, 
 \{hz54, mengnan.du\}\href{mailto:@njit.edu}{@njit.edu}
 }}
}

\begin{document}
\maketitle
\begin{abstract}
Chain of Thought (CoT) is significant in improving the reasoning abilities of large language models (LLMs). However, the correlation between the effectiveness of CoT and the length of reasoning steps in prompts remains largely unknown. To shed light on this, we have conducted several empirical experiments to explore the relations. Specifically, we design experiments that expand and compress the rationale reasoning steps within CoT demonstrations while keeping all other factors constant. We have the following key findings. First, the results indicate that lengthening the reasoning steps in prompts, even without adding new information into the prompt, considerably enhances LLMs' reasoning abilities across multiple datasets. Alternatively, shortening the reasoning steps, even while preserving the key information, significantly diminishes the reasoning abilities of models. This finding highlights the importance of the number of steps in CoT prompts and provides practical guidance to make better use of LLMs' potential in complex problem-solving scenarios. Second, we also investigated the relationship between the performance of CoT and the rationales used in demonstrations. Surprisingly, the result shows that even incorrect rationales can yield favorable outcomes if they maintain the requisite length of inference. Third, we observed that the advantages of increasing reasoning steps are task-dependent: simpler tasks require fewer steps, whereas complex tasks gain significantly from longer inference sequences. The code is available at
\href{https://github.com/MingyuJ666/The-Impact-of-Reasoning-Step-Length-on-Large-Language-Models}{%
    \nolinkurl{https://github.com/MingyuJ666/The-Impact-of-Reasoning-Step-Length-on-Large-Language-Models}
}

\end{abstract}
\renewcommand{\thefootnote}{\fnsymbol{footnote}}
\footnotetext{*Equal contribution}
\section{Introduction}
\begin{figure}[t]
    \centering
    \includegraphics[width=0.9\columnwidth]{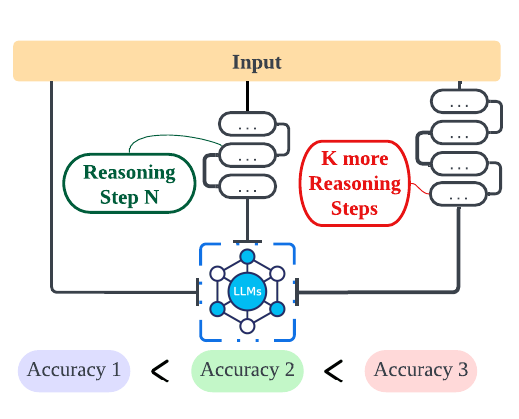}
    \caption{From left to right: zero-shot CoT, few-shot CoT, and few-shot CoT with more reasoning steps. For few-shot COT, there is a direct linear correlation between step count and accuracy.}
    \label{fig:steps-connection-with-accuracy}
\end{figure}
Today, the advent of large language models (LLMs) and their advanced prompting strategies has marked a significant progression, especially in classical NLP tasks \cite{kojima2023large,wei2022chain,shao2023synthetic,lyu2023faithful, jin2024exploring}. A key innovation among these is the Chain of Thought (CoT) prompting technique \cite{kojima2023large,wang2023selfconsistency,zhang2022automatic}, known for its efficacy in multi-step problem solving. This technique, reflecting human sequential reasoning, has shown remarkable effectiveness in various challenges, including cross-domain, length-generalization, and cross-lingual tasks. The CoT approach, with its logical, step-by-step methodology, offers crucial interpretability in complex problem-solving scenarios. Interestingly, \citeauthor{wang2023selfconsistency} found that even incorrect but coherent rationales can improve reasoning performance, highlighting the value of logical continuity \cite{wang2023selfconsistency}. Building on this, \citeauthor{fu2023complexitybased} introduced complexity-based prompting, significantly improving accuracy and setting new benchmarks \cite{fu2023complexitybased}. This research further explores the relationship between the length of reasoning steps and the accuracy of conclusions, deepening our understanding of effective problem-solving in NLP.

Despite its promising results, the research community has yet to reach a consensus on the precise mechanics of how and why CoT and its variations function effectively. This knowledge gap means that enhancing CoT performance is still a field of exploration, largely reliant on trial-and-error approaches. There still lack established systematic methodologies for improving CoT's effectiveness, leaving researchers to rely on conjecture and experimentation. This situation underscores a significant opportunity in the field: to develop a deeper, more structured understanding of CoT's inner workings. Such advancement would not only demystify the current process, but also pave the way for more reliable and efficient applications of this technique in various complex NLP tasks.

In this study, we aim to investigate the hypothesis that the reasoning steps are the most crucial element in the effectiveness of CoT prompts. This hypothesis stems from the observation that reasoning steps are a common element in both zero-shot and few-shot CoT approaches. We conduct experiments to investigate this by maintaining strict control over variables. Particularly, when incorporating new reasoning steps, we ensured that no additional knowledge was introduced. For the zero-shot experiments, we tweaked the initial prompt from ``let's think step by step'' to ``let's think step by step, you must think more steps''. Then for few-shot setting, we design experiments that expand the rationale reasoning steps within CoT demonstrations, while keeping all other factors constant.
Our first set of experiments evaluated the improvement in zero-shot and few-shot performance using Auto-CoT~\cite{zhang2022automatic} with our strategic intervention. Subsequently, we assessed the accuracy of different methods across varying step lengths. We then extended our investigation to compare the effectiveness of our strategy on different LLMs such as GPT-3.5 and GPT-4. Our findings revealed a significant correlation between the length of the reasoning chain and the capabilities of LLMs, within certain limits. Intriguingly, when we introduced misleading information into the reasoning chain, the performance still showed improvement. This highlighted a pivotal insight: The key factor appears to be the length of the thinking chain rather than its accuracy.
We have the following key findings, which we hope can help the community better improve CoT performance.
\begin{itemize}[leftmargin=*]\setlength\itemsep{-0.3em}

\item \emph{For few-shot COT, there is a direct linear correlation between step count and accuracy.} This provides a quantifiable approach for optimizing CoT prompting in complex reasoning. Specifically, lengthening the reasoning steps in prompts considerably enhances LLMs' reasoning abilities across multiple datasets. Alternatively, shortening reasoning steps, even while preserving key information, significantly diminishes the reasoning abilities of models. 

\item \emph{Even incorrect rationales can yield favorable outcomes if the required length of inference is maintained.} For example, in mathematical problems, errors in intermediate numbers have a minor impact due to their process-oriented nature.

\item \emph{The advantages of increasing reasoning steps are task-dependent}: simpler tasks necessitate fewer steps, whereas more complex tasks gain significantly from longer inference sequences. 

\item \emph{Increased reasoning steps in zero-shot CoT can also significantly improve LLM accuracy.} To validate this, we altered the initial prompt from ``Let's think step by step" to ``Let's think step by step, you must think more steps." This modification led to a noticeable enhancement in the reasoning abilities of the LLMs, particularly evident in datasets involving mathematical problems.

\end{itemize}
\section{Related Works}
In this section, we summarize two lines of literature that are most relevant to ours. 

\subsection{CoT Prompting}
The recent surge in computational power has paved the way for the rise of expansive language models. With increasing complexity, these models have unlocked emerging capabilities, notably in-context learning and CoT reasoning \cite{wei2022chain,brown2020language,schaeffer2023emergent}.

In their seminal work, Brown et al. discovered the ability of large-scale language models to leverage in-context learning (ICL)~\cite{brown2020language}. ICL strategy involves weaving input-output examples directly into the prompt, allowing ready-to-use large language models to perform impressively without the need for task-specific fine-tuning. However, despite its promise, this end-to-end methodology often falters when confronting complex reasoning challenges.

Building on this, \citeauthor{wei2022chain} demonstrated that integrating a series of logical reasoning steps into the model demonstrations, called CoT prompting, significantly refines the reasoning capabilities of large language models \cite{wei2022chain}. CoT prompting not only deepens the model's grasp of the nuanced questions and their underlying logic but also yields an articulated sequence of reasoning steps. Zhang et al.'s ``Auto-CoT" method represents a significant advancement in the field of AI reasoning \cite{zhang2022automatic}. By automating the CoT process, it addresses complex problems more effectively. 
And then Yao et al. introduced the ``Tree of Thoughts" (ToT) framework, an evolution of the Chain of Thought approach for language model inference \cite{yao2023tree}. ToT allows language models to explore different units of text as intermediate steps in problem-solving. This framework enables more deliberate decision-making by considering multiple reasoning paths. 


\subsection{Preliminary Work on Analyzing CoT}
The development and understanding of CoT reasoning in AI have evolved over time, marked by significant contributions from various researchers. Initially, Madaan and Yazdanbakhsh \cite{madaan2022text} explored the decomposition of prompts into symbols, patterns, and texts, examining the effects of CoT through counterfactual prompting. This study laid the groundwork for understanding how different components of a prompt influence AI reasoning.
Besides, several studies furthered this understanding. Tang et al. \cite{tang2023large} investigated the role of semantics in CoT reasoning, uncovering a reliance on semantic knowledge from pre-training and challenges in symbolic reasoning. Around the same time, Wang et al. focused on the impact of demonstration selection in CoT, revealing that the relevance and order of reasoning are more critical than the accuracy of reasoning chains \cite{wang2023selfconsistency}.

Theoretical perspectives also emerged recently, offering deeper insights into the mechanics of CoT. For example, Li et al. conceptualized CoT as a multi-step combinatorial function, illustrating its role in simplifying in-context learning for complex questions~\cite{li2023dissecting}. Feng et al. theoretically demonstrated the sufficiency of a fixed-size Transformer for computational tasks and dynamic planning within CoT frameworks~\cite{fu2023complexitybased}.

Further contributions in this field included those of Merrill and Sabharwal, who observed that CoT can improve reasoning abilities, with improvements scaling with the number of intermediate steps \cite{merrill2023expressive}. Wu et al. employed gradient-based feature attribution methods to assess the robustness of CoT against question variations and perturbations \cite{wu2023analyzing}.

\begin{figure*}[ht]
    \centering
    \includegraphics[width=1\linewidth]{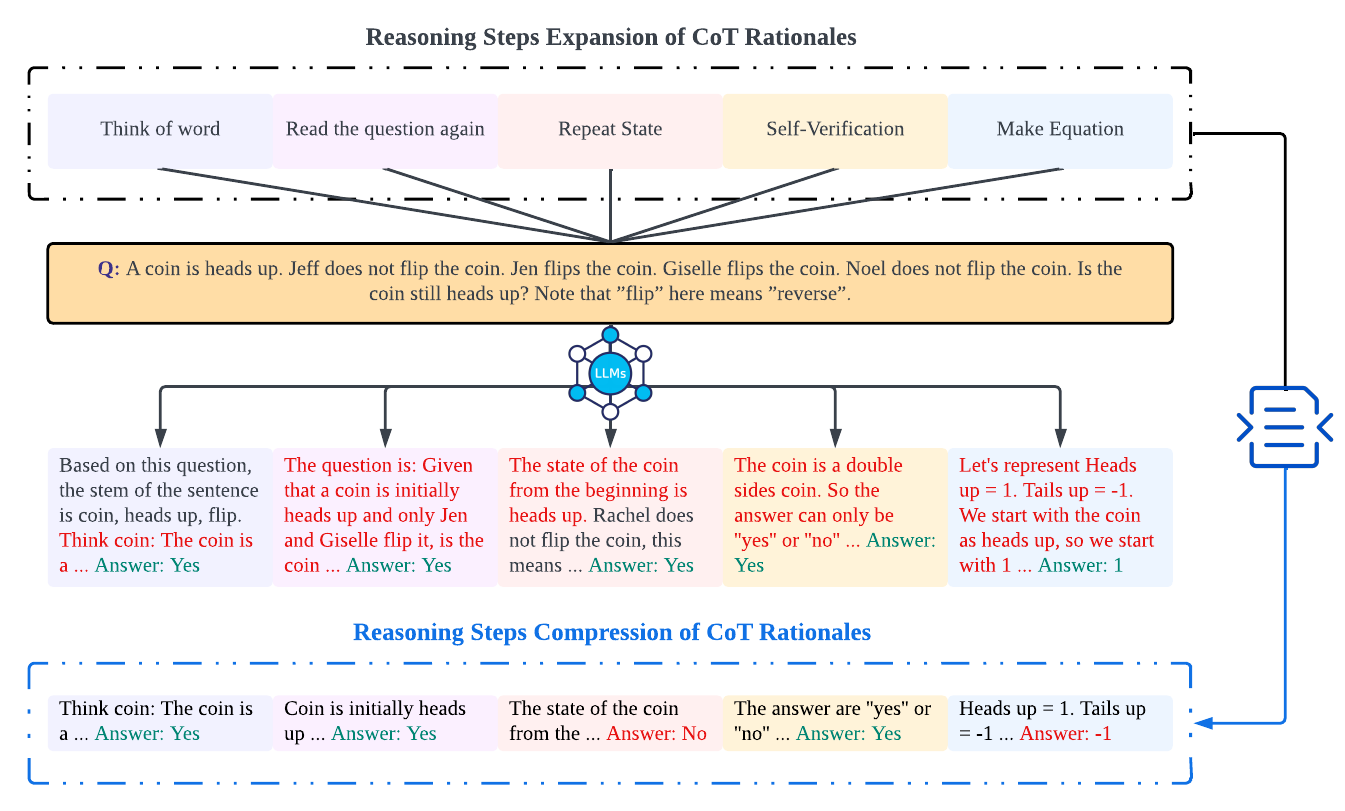}
    \caption{Increase the length of the thinking chain through the method in the figure, and compress the thinking chain without losing information as much as possible.}
    \label{fig:intro}
\end{figure*}

\section{Analyzing Methods}\label{sec:methods}
In this section, we propose an analysis to examine the relationship between the reasoning steps and the chain-of-thought (CoT) prompting performance. Our central hypothesis is that the reasoning steps are the most critical component of CoT prompts, enabling language models to apply more logical reasoning when generating responses.
To test this, we design experiments that expand and compress the rationale reasoning steps within CoT demonstrations, while keeping all other factors constant. Specifically, we systematically vary only the number of reasoning steps, without introducing new reasoning content or removing existing reasoning content. We evaluate both zero-shot and few-shot CoT prompts in the following sections. The overall experimental procedure is illustrated in ~\autoref{fig:intro}. Through this controlled analysis, we aim to shed light on how CoT influences the LLM's ability to produce logically sound responses. 

\subsection{Preliminary}
Zero-Shot-CoT \cite{kojima2023large} is a template-based zero-shot prompt for chain-of-thought reasoning. The core idea is to add \textit{``Let's think step by step''} or other similar text. Compared to Zero-Shot-CoT, Few-Shot-CoT provides more examples in the prompt for chain-of-thought reasoning. Among them, Manual-CoT \cite{wei2022chain}, Auto-CoT \cite{zhang2022automatic} are two popular methods.

\noindent
\textbf{Manual-CoT}: Manual-CoT prompting relies on a few manually designed demonstrations, each composed of a question and a reasoning chain leading to an answer, to improve language models' reasoning performance.

\noindent
\textbf{Auto-CoT}: Auto-CoT eliminates the need for manual demonstration design by automatically constructing demonstrations through clustering test questions to select diverse examples and generating reasoning chains using the language model's own zero-shot reasoning capability.

\subsection{Analyzing Zero-shot CoT}
\begin{table*}[t]
\small
  \caption{Case of Think About The Word, select a few words and ask the model to think about the meaning in text.}
  \label{Think}
  \begin{tabularx}{\textwidth}{X}
    \toprule
    \multicolumn{1}{c}{\textbf{Think About The Word}} \\
    \midrule
    \textbf{Prompt}: \\
    Q: Could someone in Tokyo take a taxi to the Metropolitan Museum of Art?\\
    A: Let's think step by step. The stem of the sentence is Tokyo, take a taxi, Metropolitan Museum of Art. \textcolor{red}{Think about Tokyo... Think about taking a taxi... Think about Metropolitan Museum of Art...} Inference: Tokyo is in Japan and the Metropolitan Museum of Art is in New York. The two places are separated by the sea, so you can't take a taxi there. The answer is no. \\
    \midrule
    Q: Could a chipmunk fit 100 chocolate chips in his mouth?\\
    \textbf{Answer}:\\
    A: Based on this question, the stem of the sentence is chipmunk, fit, 100 chocolate chips, mouth. \textcolor{red}{Think about chipmunk... Think about fit... Think about chocolate chips...} Inference: Chipmunks are small animals with relatively small mouths, so it's unlikely that a chipmunk would be able to fit 100 chocolate chips in its mouth at once. The answer is no.\\
    \textbf{Prediction}: no\\
    \textbf{Ground Truth}: no\\
    \bottomrule
  \end{tabularx}
\end{table*}

In the zero-shot scenario, we modified the initial prompt from ``Let's think step by step" to ``Let's think step by step, you must think more steps." This change was implemented because, unlike the Few-shot CoT context, we cannot introduce additional reasoning steps in the example. By altering the initial prompt, we guide the LLM to engage in more extensive thinking. This approach is crucial as it enhances the model's accuracy without the need for incremental training or additional example-driven adjustments typical in few-shot CoT scenarios. This refined strategy ensures a more comprehensive and detailed reasoning process, thereby significantly improving the model's performance in zero-shot settings.

\subsection{Analyzing Few-shot CoT}
In this section, we aim to modify the reasoning chains within CoT rationales, either by adding or compressing reasoning steps. The goal is to examine how changes in reasoning structure influence LLM decision-making. During rationale expansion, we will avoid introducing any new task-relevant information. This isolates reasoning steps as the only variable under study.

To this end, we plan to investigate the following strategies to expand the reasoning steps for different LLM applications.
There are usually fixed patterns in the way people think about a problem, for example, by repeating the question over and over again to gain a deeper understanding, by creating mathematical equations to reduce the burden on memory, by analyzing the meaning of words in the question to help understand the topic, by summarizing the current state to simplify the description of the topic. 
Based on the inspiration of Zero-Shot-CoT and Auto-CoT, we expected the process of CoT to become a standardized pattern, and lead to the right result by restriction on the direction of CoT thinking in the prompt section.
The core of our approach is to simulate the process of human thinking and reshape the chain of thought. We give five general prompt strategies in \autoref{wide_table1} in the Appendix.

\begin{itemize}[leftmargin=*]\setlength\itemsep{-0.3em}
\item 
\textbf{Think About The Word}: This strategy is to ask the model to interpret the word and rebuild the knowledge base. Typically a word has multiple different meanings, and the effect of this is to get the model to think outside the box and reinterpret the words in the problem based on the generated interpretations. This process does not introduce new information. In the prompt, we give examples of the words that the model is thinking about, and the model automatically picks words for this process based on the new question.

\item 
\textbf{Read the question again}: Read the questions repeatedly to reduce the interference of other texts on the chain of thought. In short, we let the model remember the questions.

\item 
\textbf{Repeat State}: Similar to repeated readings, we include a small summary of the current state after a long chain of reasoning, aiming to help the model simplify its memory and reduce the interference of other texts in the CoT.

\item 
\textbf{Self-Verification}: Humans will check if their answers are correct when answering questions. Therefore, before the model gets the answer, we add a self-verification process to judge whether the answer is reasonable based on some basic information.

\item 
\textbf{Make Equation}: For mathematical problems, Make Equations can help humans summarize and simplify memory. And for some problems that require the assumption of an unknown number $x$, establishing an equation is an essential process. We simulated this process and let the model try to make equations in mathematical problems.
\end{itemize}

Overall, our prompt strategies all saw corresponding patterns in the model's responses. We give an example in ~\autoref{Think}, examples of the other four strategies can be found in the appendix. In Section 4 we will perform a quantitative analysis to validate the effectiveness of our strategies. We assume that each additional strategy is equivalent to increasing the reasoning step length by one.

\section{Experimental Results}

\begin{table*}[!h]
\caption{Comparison of accuracy of our method with four baselines on eight datasets}
\centering
\resizebox{1.0\textwidth}{!}{
\begin{tabular}{lcccccccc}
\hline
\multirow{2}{*}{Models} & \multicolumn{5}{c}{Arithmetic} & \multicolumn{1}{c}{Commonsense} & \multicolumn{2}{c}{Symbolic} \\
\cmidrule(lr){2-6}
\cmidrule(lr){7-7}
\cmidrule(lr){8-9}

& MultiArith & GSM8K & AQuA  & SingleEq & SVAMP & Strategy & Letter & Coin \\
\hline
Zero-Shot & 40$\pm$1.0 & 30.4$\pm$1.7 & 29.9$\pm$1.8 & 82.7$\pm$1.3 & 56$\pm$1.0 & 59$\pm$1.0 & 43$\pm$1.0 & 79.8$\pm$1.2 \\
Zero-Shot-CoT & 91.5$\pm$1.2 & 64.1$\pm$1.1 & 55.6$\pm$1.3 & 87.43$\pm$0.25 & 79.99$\pm$1.41 & 58.34$\pm$1.56 & 69$\pm$1.0 & 93$\pm$1.0  \\
Manual-CoT & 93.5$\pm$0.1 & 64.7$\pm$1.5 & 55$\pm$1.0 & 92.1$\pm$0.2 & 82.3$\pm$0.3 & 65.3$\pm$1.4 & 75$\pm$0.0 & 92.7$\pm$0.1 \\
Auto-CoT & 94$\pm$0.0 & 65.8$\pm$0.9 & 65$\pm$0.0 & 92$\pm$0.0 & 81.9$\pm$0.3 & 65.3$\pm$0.5 & 73.5$\pm$0.2 & 93$\pm$0.0 \\
\hline 
Must Think More Step (Zero-Shot-CoT)  & 95.2$\pm$0.3 & 76.1$\pm$0.1 & 62.11$\pm$0.24& 87.43$\pm$0.16 & 79.99$\pm$0.18 & 72.6$\pm$0.21 & 69$\pm$0.0 & 97$\pm$0.0 \\
Add Reasoning Step (Manual-CoT) & 97$\pm$0.0 & 70.1$\pm$0.3 & 62.5$\pm$0.5 & 88.97$\pm$0.27 & 85.2$\pm$0.2 & 68.86$\pm$0.27 & 77.8$\pm$0.4 & 97.3$\pm$0.1 \\
Add Reasoning Step (Auto-CoT) & 97.2$\pm$0.1 & 78.8$\pm$0.2 & 64.03$\pm$0.36 & 92.71$\pm$0.14 & 83.7$\pm$0.2 & 70.26$\pm$0.19 & 71.2 $\pm$0.5& 99$\pm$0.0 \\
\hline
\end{tabular}%
}
\label{tab:main-results}
\end{table*}
We conduct experiments to answer the following research questions: RO1: What is the relationship of rational reasoning steps in demonstrations with CoT performance? (Section 4.2) RO2: Can we confirm that the reasoning steps are the only factor that affects LLM performance? (Section 4.3) RO3: Will compressing reasoning steps in few-shot demonstrations hurt LLM performance? (Section 4.4), RO4: Can we observe the scaling phenomenon, i.e., the required reasoning steps be related to the size of the LLMs? (Section 4.5) and RO5: What is the influence of questions in rationale towards the LLM reasoning ability? (Section 4.6)

\begin{table*}[!h]
\caption{The Case of Wrong Prompt, change one of the step in the chain of thought and preserve overall coherence}
\resizebox{1.0\textwidth}{!}{
\label{tab:case-wrong}
\begin{tabular}{|l|l|c|c|}
\hline
\multicolumn{1}{|c|}{Original Prompt}                                & \multicolumn{1}{|c|}{Wrong Prompt}  \\ \hline
\begin{tabular}[c]{@{}l@{}}\textbf{Q:} Joan has 10 books. Tom has 38 books. \\How many books do they have?\\
\textbf{Rationale:} Let’s think step by step. Joan has 10 books.\\
Tom has 38 books. we have equation books = 10 + 38 = 48. \\
They have 10 + 38 = 48 books together.\\ \textbf{Q:} Megan had 217 markers. Robert gave her 109 more markers.\\
How many markers does Megan have altogether?\end{tabular}  & {\begin{tabular}[c]{@{}l@{}}\textbf{Q:} Joan has 10 books. Tom has 38 books. \\How many books do they have?\\
\textbf{Rationale:} Let’s think step by step. Joan has 10 books.\\
Tom has 38 books. we have equation books = 10 + \textcolor{red}{8} = 48. \\
They have 10 + 38 = 48 books together.\\
\textbf{Q:} Megan had 217 markers. Robert gave her 109 more markers.\\
How many markers does Megan have altogether?\end{tabular}}  \\ \hline
\end{tabular}}
\end{table*}
\begin{figure*}[!h]

	\subfigbottomskip=2pt 

	\subfigcapskip=-2pt 
	\subfigure[MultiArith]{\includegraphics[width=0.24\linewidth]{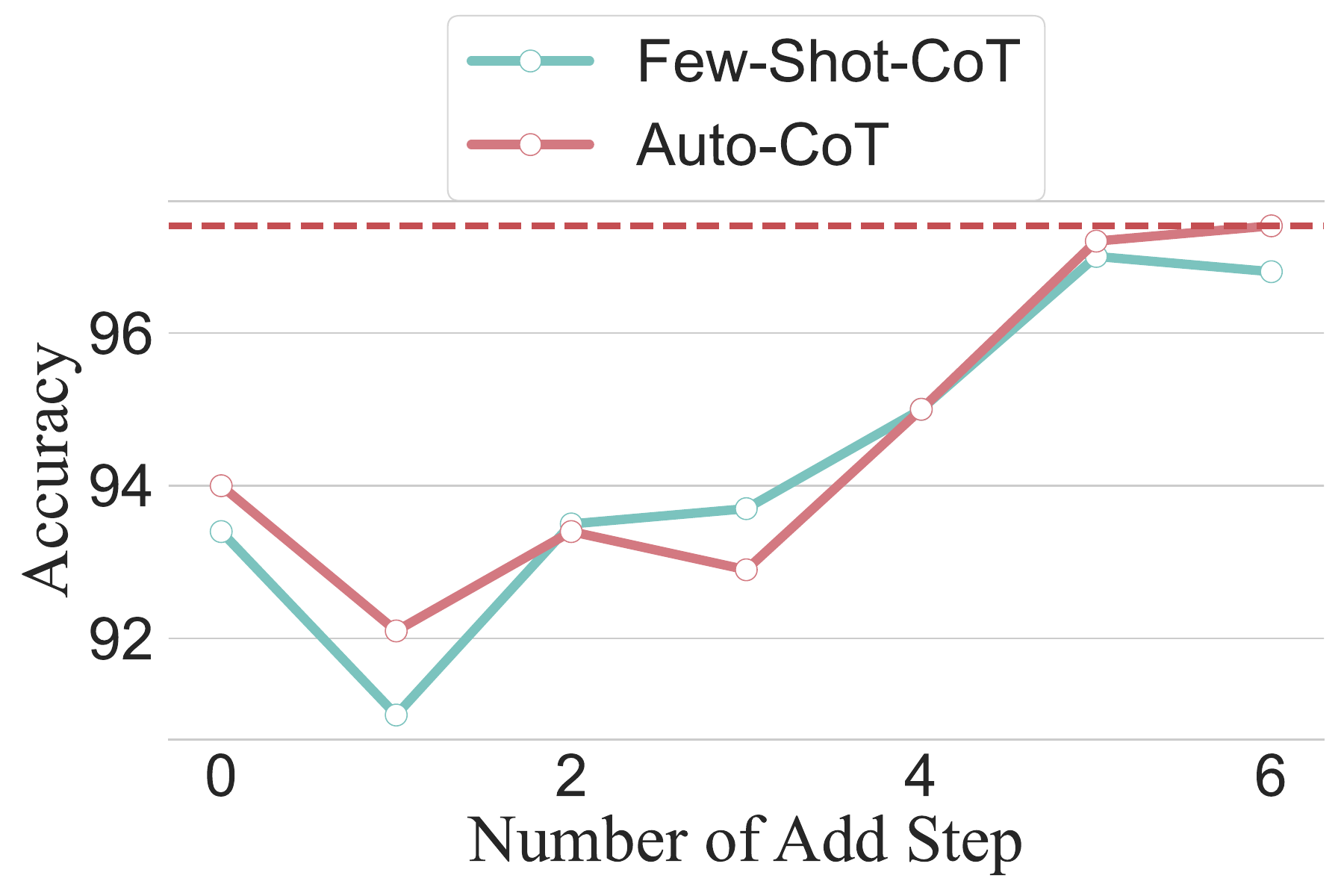}}
	\subfigure[GSM8K]{\includegraphics[width=0.24\linewidth]{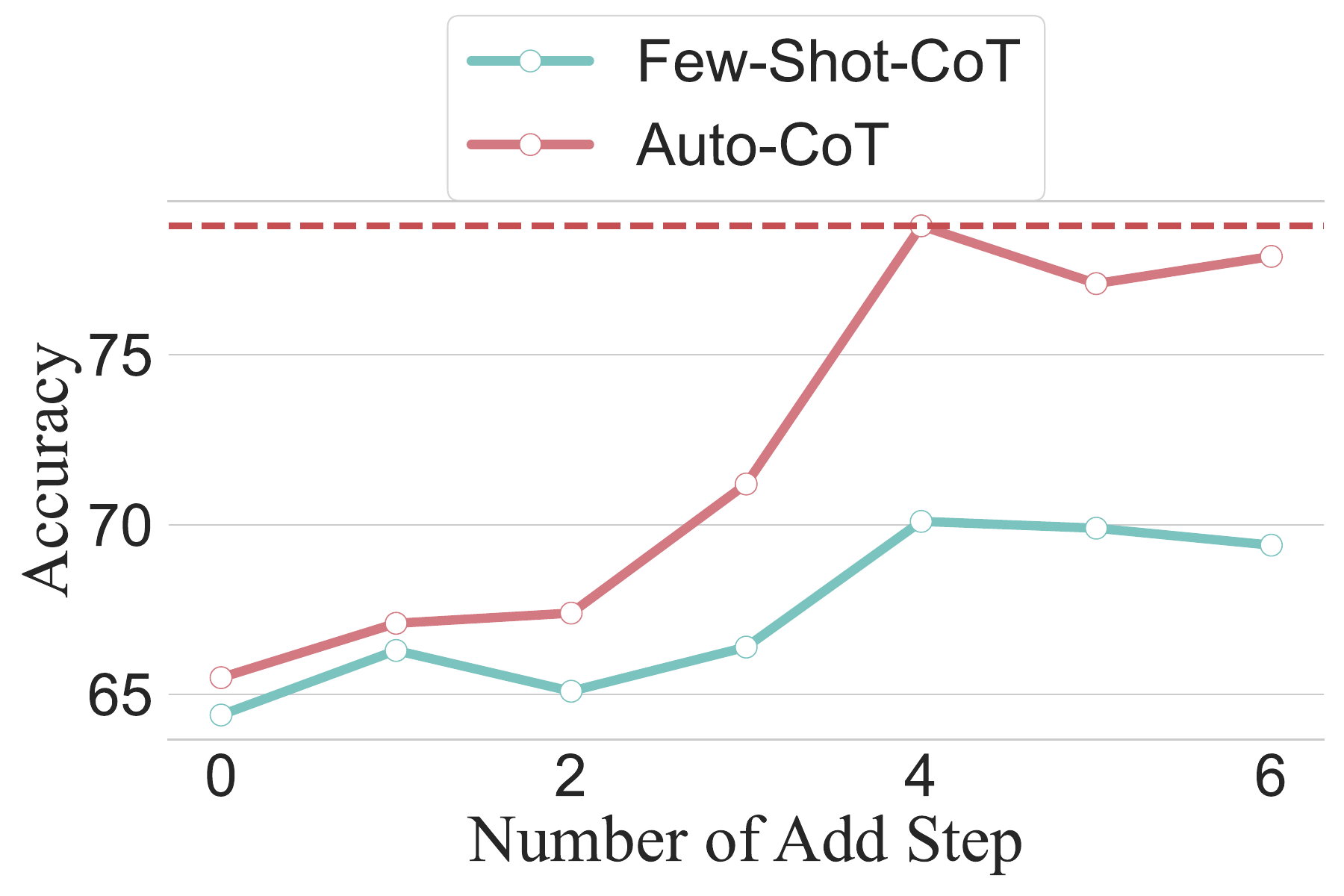}}
	  \subfigure[AQuA]{\includegraphics[width=0.24\linewidth]{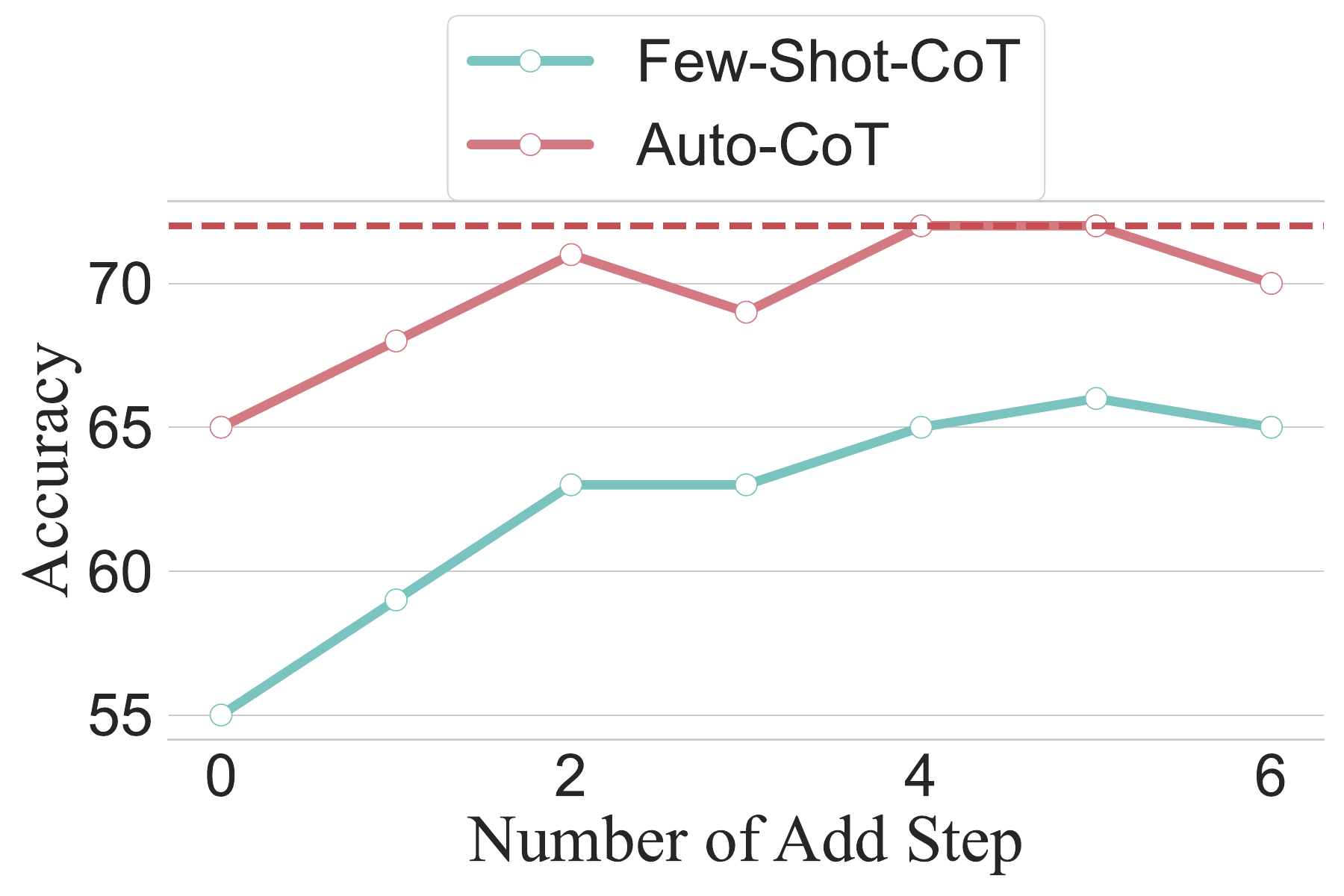}}
        \subfigure[SingleEq]{\includegraphics[width=0.24\linewidth]{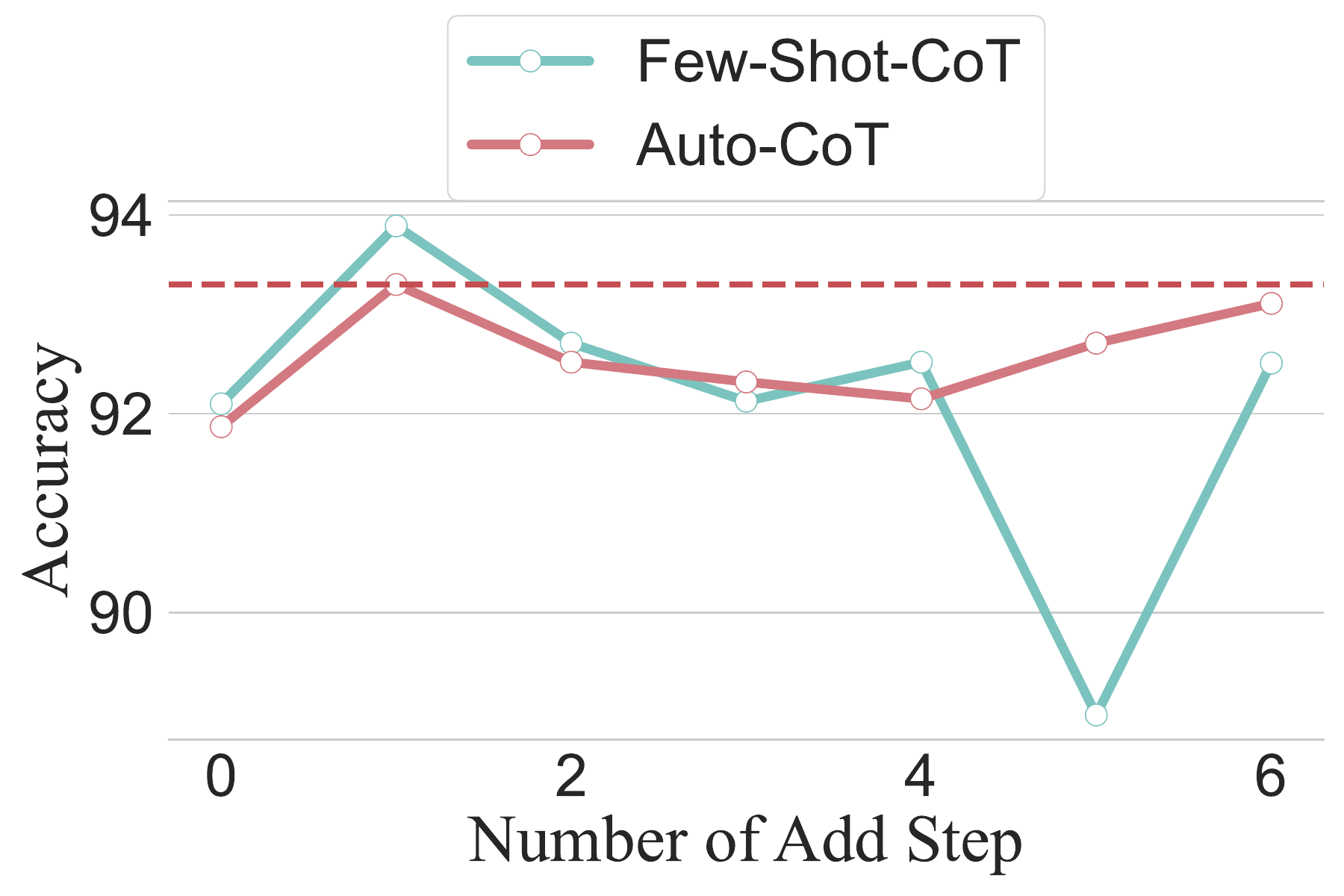}}
   
	\subfigure[SAVMP]{\includegraphics[width=0.24\linewidth]{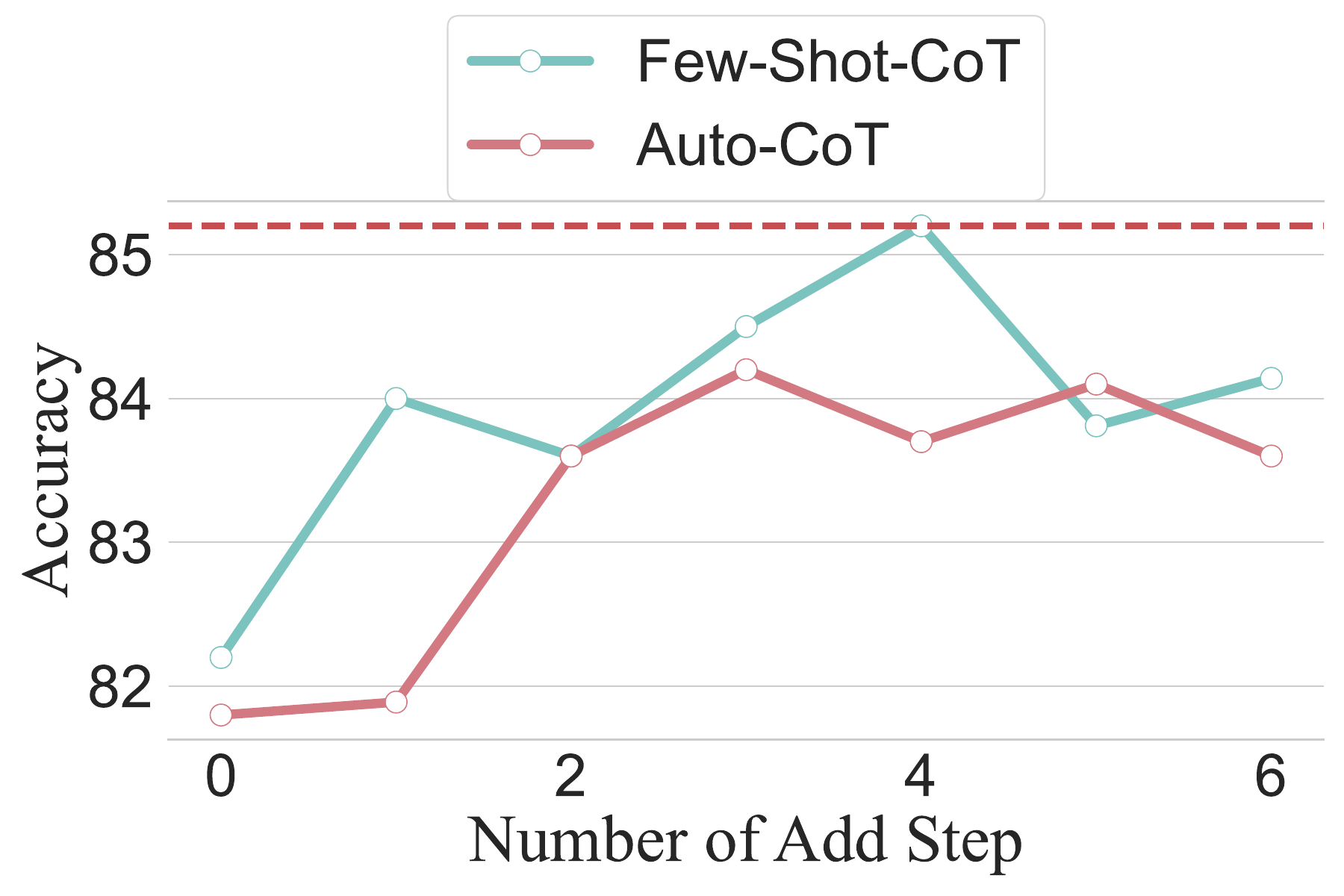}}
	\subfigure[strategyqa]{\includegraphics[width=0.24\linewidth]{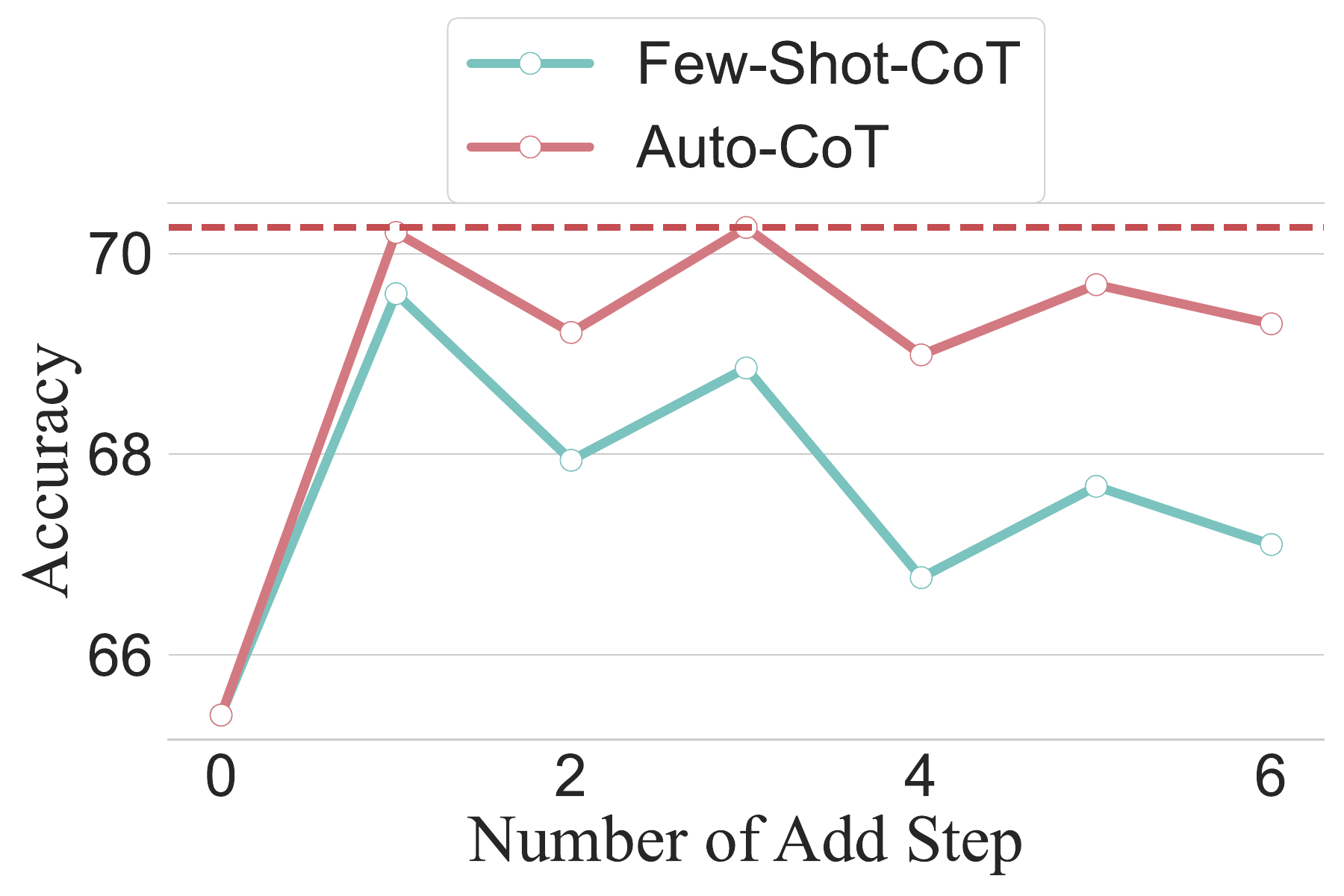}}
	\subfigure[Letter]{\includegraphics[width=0.24\linewidth]{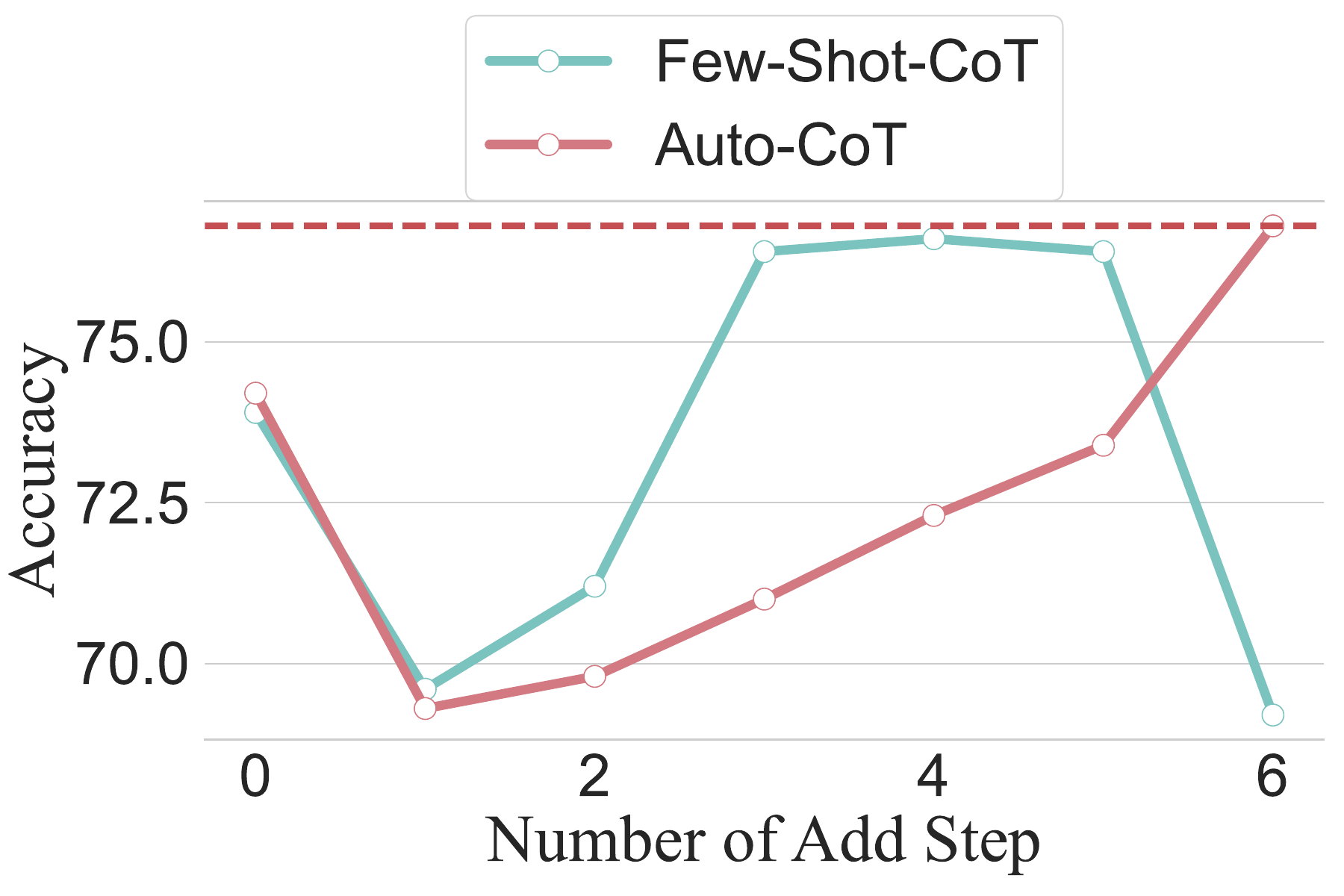}}
	\subfigure[Coin]{\includegraphics[width=0.24\linewidth]{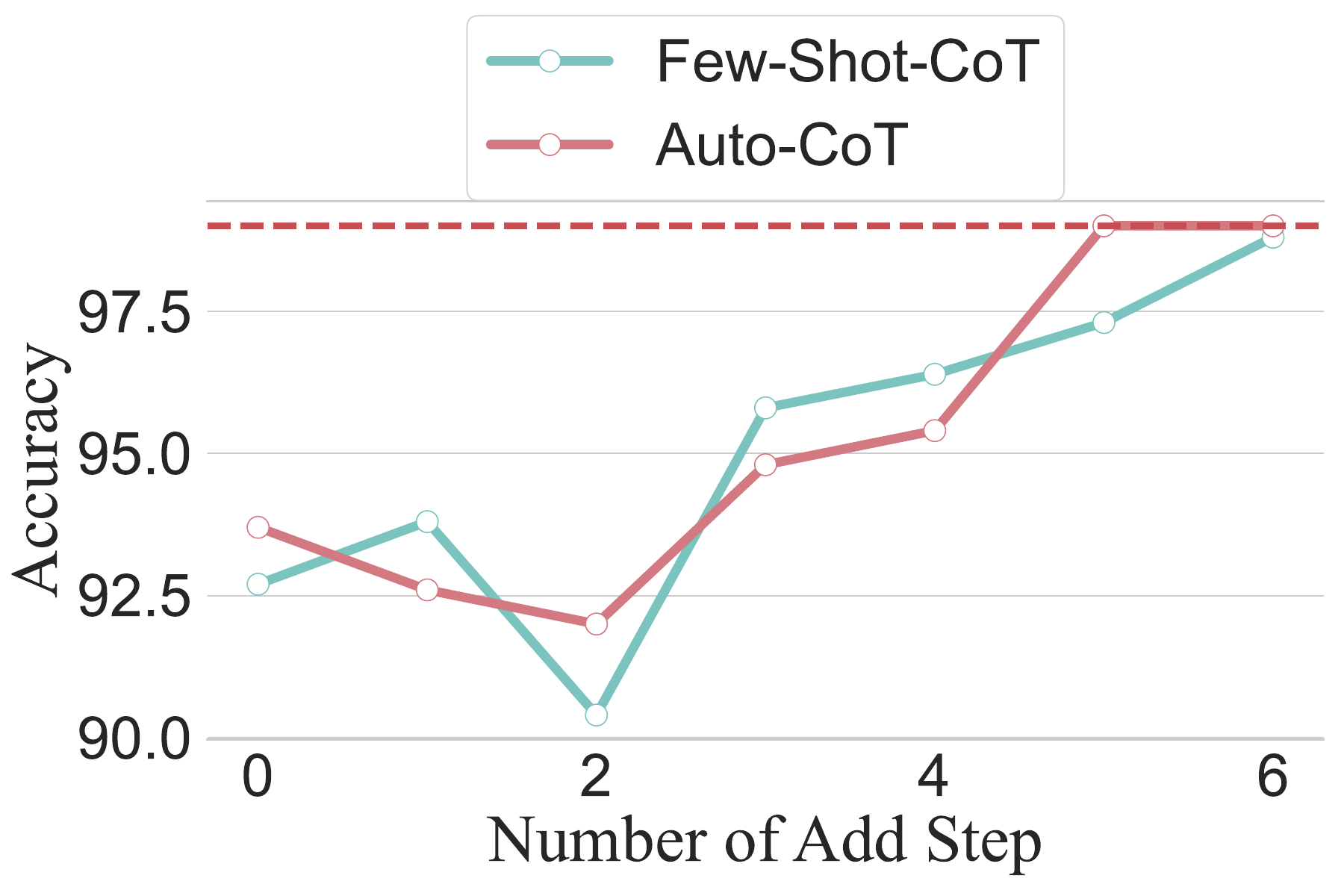}}
\caption{Linear Relationship Between Step Quantity and Accuracy}
\label{figure1}
\end{figure*}

\subsection{Experimental Setups}
\label{section4.1}
We introduce the general experimental setups.

\noindent
\textbf{Datasets and Models}.\, We evaluate our proposal on eight dataset (MultiArith \cite{roy2015solving}, GSM8K \cite{cobbe2021training}, AQuA \cite{ling2017program}, SingleEq \cite{koncel2015parsing}, SAVMP \cite{patel2021nlp}, Letter \cite{wei2022chain}, Coin \cite{wei2022chain}, Strategyqa \cite{geva2021did}). We employ three models to validate the performance of our proposed method, which are: text-davinci-002~\cite{brown2020language}, GPT-3.5-turbo-1106 \cite{ouyang2022training}, GPT-4 \cite{openai2023gpt}.  All models are accessed via the OpenAI API key.

\vspace{2pt}
\noindent
\textbf{Prompt}.\, We have shown the proposed process pipeline in Section~3 Analyzing Methods. The experimental part follows the same approach.

\vspace{2pt}
\noindent
\textbf{Baselines}.\, We compare our methods with four baseline methods: Zero-Shot \cite{kojima2023large}, Zero-Shot-CoT \cite{kojima2023large}, Manual-CoT \cite{wei2022chain}, Auto-CoT \cite{zhang2022automatic}. The results are in the ~\autoref{tab:main-results}. 

\vspace{2pt}
\noindent
\textbf{Evaluation Metrics}.\,
Accuracy is used to assess a model's ability on classification tasks and is commonly used for multichoice and yes/no tests:
$
\text{Accuracy} = {N_{\text{correct}}}/{N_{\text{total}}}.
$

\vspace{2pt}
\noindent
\textbf{Implementation Details}:

\begin{itemize}[leftmargin=*]\setlength\itemsep{-0.3em}
\item Add reasoning steps: This process deploys GPT-4 to modify the Zero-Shot-CoT prompt demo generated by ``let's think step by step" to contain the five reasoning steps we mentioned in Section~\ref{sec:methods} so that we can define the number of steps and the types of steps to be contained in the demo. We then input the demo as a prompt. We completed the following experiment with this approach.

\item Reasoning-Step Compression: In the expressing experiment, we focused on executing a compression attack on the rationale inference chain within the few-shot CoT. The process involved randomly selecting two consecutive sentences and employing GPT-4 to effectively merge them. We then input the prompt: ``Please compress the following two sentences without losing any information, and make them as concise as possible''. This method was designed to implement a targeted compression on the reasoning chain.

\item 
Answer cleaning: We follow the structure proposed by Zero-Shot-CoT to select the final answer. After the model response output is obtained, this structure selects only part of the answer that first satisfies the answer format.

\end{itemize}


\subsection{Relationship Between Steps and Accuracy} 
\label{section4.2}
\autoref{tab:main-results} compares accuracy on eight datasets from three categories of reasoning tasks using GPT-3.5-turbo-1106. All results are averaged over three random runs. Our SOTA results are based on experimental results from the optimal performance step for each data set. Our zero-shot CoT is based on Section 2.1, and the Add Reasoning Step (Manual-CoT), and Add Reasoning Step (Auto-CoT) is based on Section 2.2.\\

Benefiting from the fact that we have standardized the thought chain process, it is possible to quantify the increase in accuracy due to the increased steps in rationales of COT demonstrations. We conducted experiments to answer the RO1: What is the relationship of rational reason-ing steps in demonstrations with CoT performance? This experiment is completed with GPT-3.5-turbo-1106, and the results are given in ~\autoref{figure1}. We found that LLM reasoning ability improves in all datasets during an effective CoT process, i.e. with the addition of up to six steps of additional thought processes. In other words, we found a certain linear relationship between accuracy and CoT complexity.

\subsection{Effect of Prompt with Wrong Answer}
\label{section4.3}
\begin{figure}[t]
    \centering
    \includegraphics[width=1\linewidth]{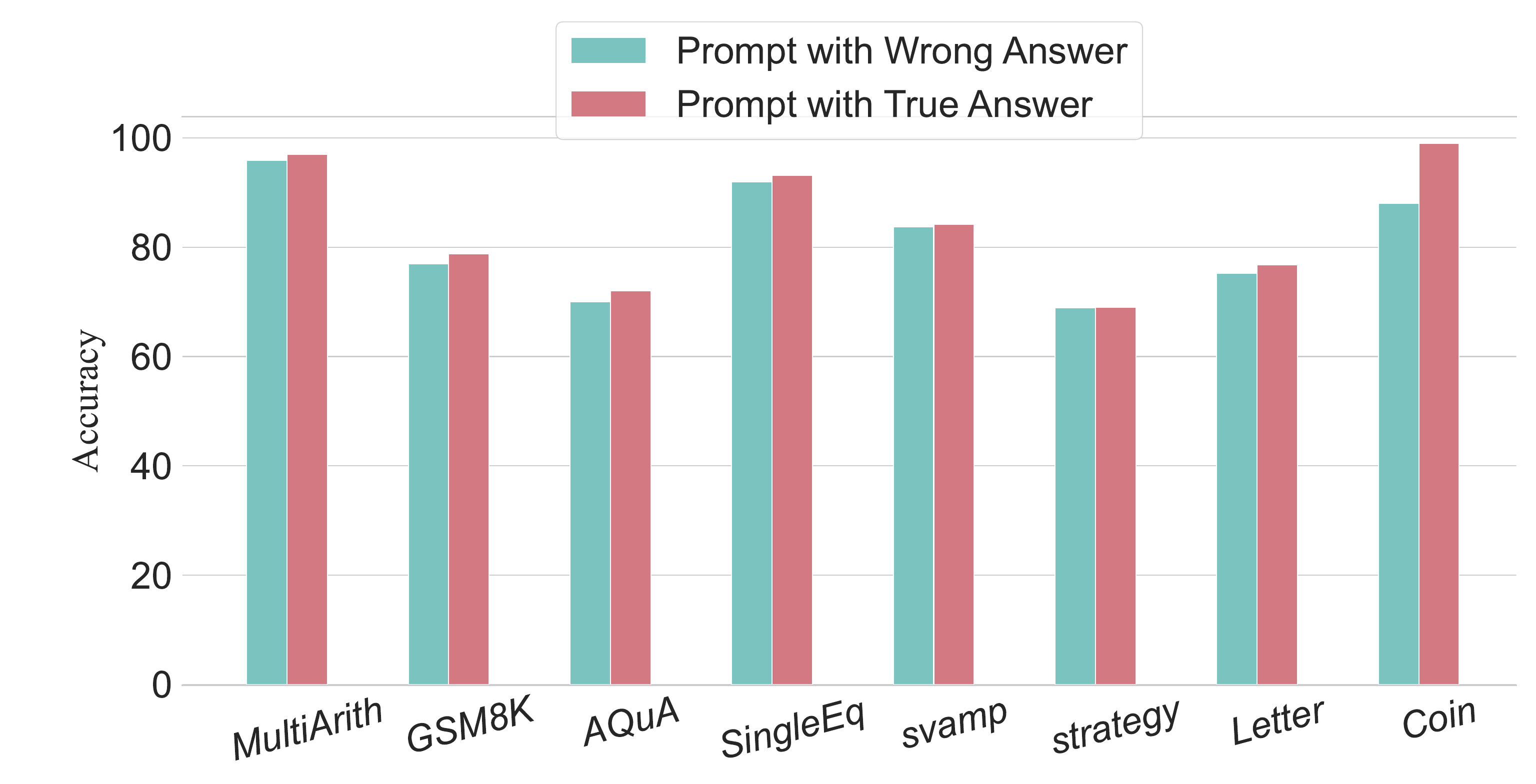}
\caption{Compare the accuracy of the prompt with the true answer and the prompt with the wrong answer}
\label{fig:wrong-prompt}
\end{figure}

To answer RO2: Are reasoning steps the only factor that affects LLM performance? We made the following attempt. Change a step in the prompt to an incorrect answer to see if it affects the chain of thought. So, for this experiment, we change all the prompts to carry one error. For a concrete example, check the ~\autoref{tab:case-wrong}. For Arithmetic-type questions, even if there is a deviation in one of the results of the prompt, the effect on the chain of thought in the reasoning process is minimal, so we believe that the large language model learns more about the chain of thought patterns in the prompt than the single computation when solving Arithmetic-type problems. For logic problems similar to those in the Coin dataset, a deviation in one of the prompt's results often brings about the fragmentation of the entire chain of thought. We completed this experiment with GPT-3.5-turbo-1106 and guaranteed performance based on the optimal number of steps for each data set derived from the previous experiment. The results are shown in ~\autoref{fig:wrong-prompt}.

\subsection{Compressing Reasoning Steps }
\label{section4.4}
\begin{figure}[t]
    \centering
    \includegraphics[width=1\linewidth]{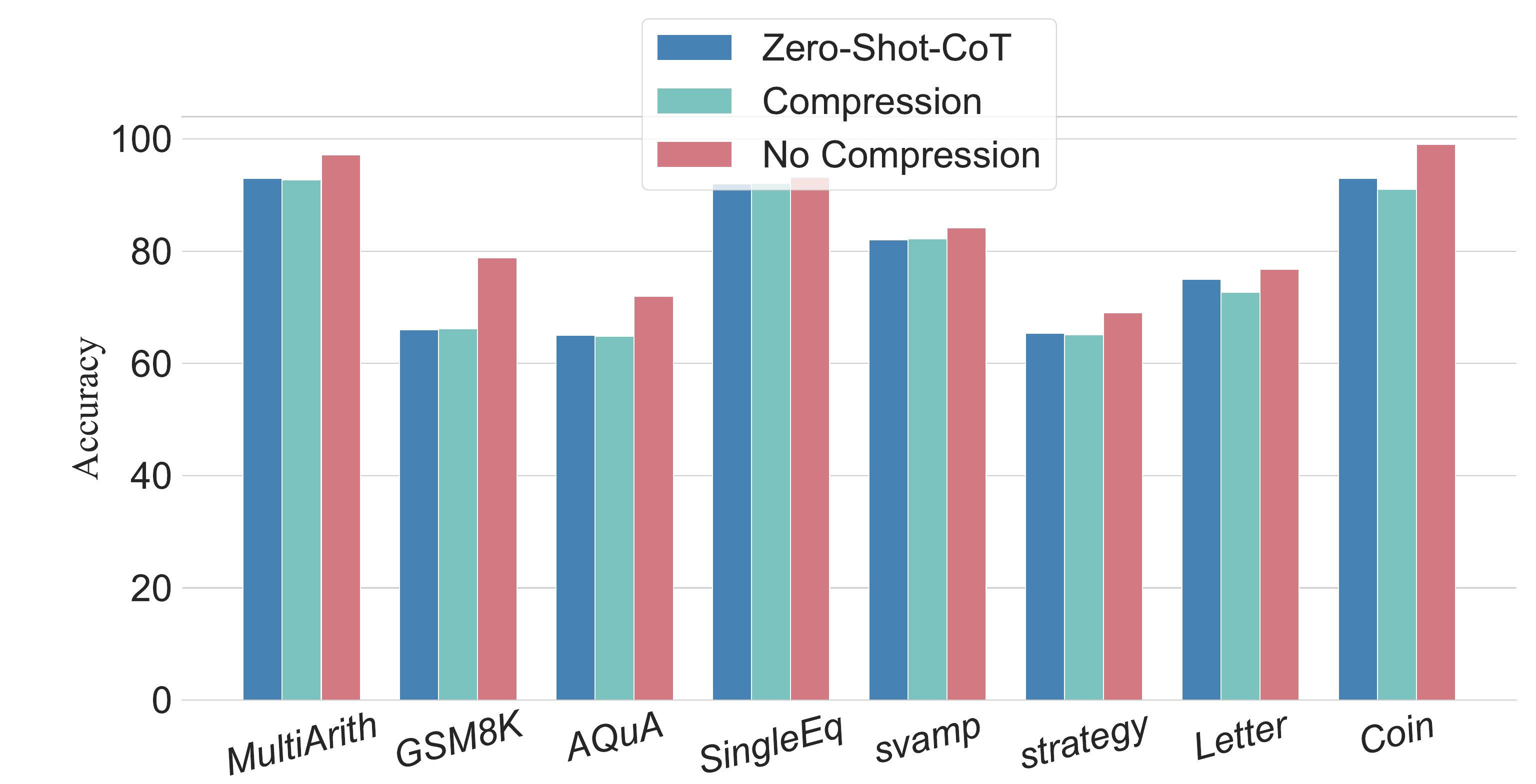}
   \caption{Compare the accuracy of the prompt with Compression and the prompt with No Compression}
    \label{fig:Compression}
\end{figure}

\begin{table*}[!h]
\small
\caption{Making deliberate alterations to sample questions}
\resizebox{\textwidth}{!}{
\label{tab:case-wrong question}
\begin{tabular}{|m{\textwidth}|}
\hline
\textbf{Original Prompt} \\ \hline
\textbf{Q:} Wendy uploaded 45 pictures to Facebook. She put 27 pics into one album and put the rest into 9 different albums. How many pictures were in each album? \\
\textbf{Rationale:} ``Let's think step by step. First, Wendy uploaded 45 pictures in total. Second, Wendy put 27 pictures into one album. That means that Wendy put the remaining 18 pictures into 9 different albums. Each album would have 2 pictures." \\
\textbf{Pred\_ans:} ``2" \\
\textbf{Gold\_ans:} ``2" \\ \hline
\textbf{Making deliberate alterations} \\ \hline
\textbf{Q:} \textcolor{red}{Wendy uploaded 66 pictures to Facebook. She put 89 pics into one album and put the rest into 7 different albums. How many pictures were in each album?} \\
\textbf{Rationale:} ``Let's think step by step. First, Wendy uploaded 54 pictures in total. Second, Wendy put 27 pictures into one album. That means that Wendy put the remaining 12 pictures into 6 different albums. Each album would have 7 pictures." \\
\textbf{Pred\_ans:} ``2" \\
\textbf{Gold\_ans:} ``2" \\ \hline
\end{tabular}}
\end{table*}
In previous sections, we have demonstrated that increasing reasoning steps could improve the LLM reasoning accuracy. In this section, our aim is to answer RO3: Will compressing reasoning steps in few-shot demonstrations hurt LLM performance?
To this end, we conduct the reasoning steps compression experiment, and we employed the technique outlined in the experimental setup to condense the reasoning process in examples from both the baseline automated chain of thought (Auto-CoT) and the few-shot chain of thought (Few-Shot-CoT), aiming to reduce their number of inference steps. The result is shown in ~\autoref{fig:Compression}. It revealed a notable decline in performance, which regressed to a level essentially equivalent to that achieved by the zero-shot method. It further indicates that \emph{increasing COT rationale reasoning steps could improve COT performance and the vice versa.}

\subsection{Performance on Different Size Models}
\label{section4.5}

\begin{figure}[t]
    \centering
    \includegraphics[width=0.75\linewidth]{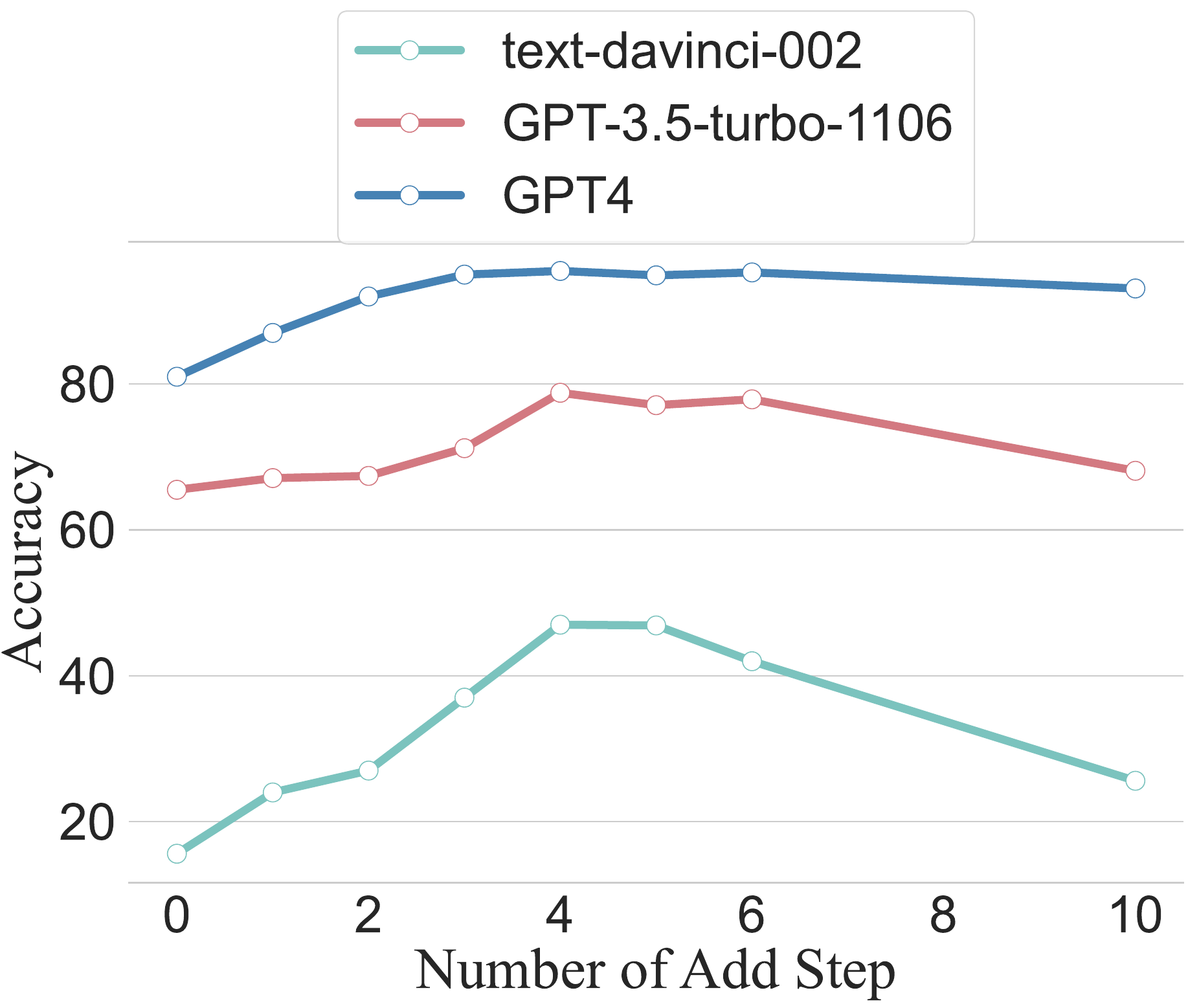}
    \caption{Comparing the accuracy with different size model on dataset GSM8K.}
    \label{diff_model}
\end{figure}

In this chapter, our focus is to answer RO4: Can we observe the scaling phenomenon, i.e., the required reasoning steps be related to the size of the LLMs? We examine the average number of inference steps utilized in various models, including text-davinci-002, GPT-3.5-turbo-1106, and GPT-4. We conducted experiments on GSM8K calculating the average inference step for each model to reach peak performance. This dataset has the largest performance difference with text-davinci-002, GPT-3.5-turbo-1106, and GPT-4 out of our 8 datasets. It can be observed that the model with the worst initial performance, text-davinci-002, our strategy has the highest boosting effect. For the model with the best initial performance, GPT-4, has the highest tolerance to our strategy (no performance decreases). We show the result in ~\autoref{diff_model}.

\subsection{Influence of Questions in CoT Examples}
\label{section4.6}

In our case study, we aim to answer RO5: What is the influence of questions in rationale towards the LLM reasoning ability? We want to explore whether changing the reasoning in CoT will affect CoT performance. Since we are mainly studying the impact of the inference step on performance, we need to confirm that the question has no impact on performance. So we have chosen two datasets and two CoT methods (auto-CoT and few-shot-CoT) for this investigation: MultiArith \cite{roy2015solving} and GSM8K \cite{cobbe2021training} in GPT-3.5-turbo-1106. Our experimental approach involves making deliberate alterations to sample questions within these mathematical datasets, such as varying the content of the questions in ~\autoref{tab:case-wrong question}. Remarkably, initial observations indicate that these modifications have only minimally impacted performance like ~\autoref{tab:case1}.

This preliminary finding suggests that the length of the steps involved in the reasoning process, rather than the nature of the questions themselves, predominantly influences the reasoning capabilities of large-scale models. 

\begin{table}[t]
\small
\centering
\caption{Accuracy comparison of models on different datasets}
\begin{tabular}{lcc}
\toprule
Models & MultiArith & GSM8K \\
\midrule
Zero-Shot & 40$\pm$1.0 & 30.4$\pm$1.7 \\
Zero-Shot-CoT & 91.5$\pm$1.2 & 64.1$\pm$1.1 \\
Manual-CoT & 93.5$\pm$0.1 & 64.7$\pm$1.5 \\
Auto-CoT & 94$\pm$0.0 & 65.8$\pm$0.9 \\
\hline 
Changing Question \\(Manual-CoT) & 92.9$\pm$0.1 & 62.1$\pm$1.7 \\
Changing Question \\(Auto-CoT) & 92.5$\pm$0.1 & 63.5$\pm$1.0 \\
\hline
Add Reasoning Step \\(Manual-CoT) & 97$\pm$0.0 & 70.1$\pm$0.3 \\
Add Reasoning Step \\(Auto-CoT) & 97.2$\pm$0.1 & 78.8$\pm$0.2 \\
\hline
Add Reasoning Step \\and Changing Question\\(Manual-CoT) & 96.6$\pm$0.1 & 69.6$\pm$0.2 \\
Add Reasoning Step \\and Changing Question\\(Auto-CoT) & 95.7$\pm$0.1 & 75.2$\pm$0.2 \\
\bottomrule
\end{tabular}
\label{tab:case1}
\end{table}

\section{Conclusions and Future Work}
In this work, we make a critical contribution to understanding and optimizing CoT in LLMs, especially in the realm of complex reasoning tasks. Our extensive research on the CoT technique in natural language processing, particularly with large language models like GPT-3, GPT-3.5, and GPT-4, has led to key insights. We found a notable correlation between the length of the reasoning chain and the performance of these models. Interestingly, longer reasoning chains improve model performance, even when they contain misleading information. This suggests that the chain's length is more crucial than its factual accuracy for effective problem-solving. These findings provide valuable guidance for refining CoT strategies, highlighting the significance of reasoning length in complex NLP tasks.

Our next step is to analyze the long and short reasoning steps of LLM inference via explaindetermine
Our objective is to ascertain whether longer inferential steps correlate with broader neuronal engagement. To illustrate this, we intend to use visualization techniques to analyze activation patterns between long and short reasoning steps. 
\section{Limitation}
In this work, we provide an experimental analysis of how CoT works. We focus specifically on manipulating the reasoning steps in CoT prompts and measuring the impact on model performance. However, we did not deeply analyze the underlying mechanisms behind why increasing reasoning steps improves performance. This includes either theoretical analysis or explainability analysis to analyze the internal workings of LLMs, and further investigation could provide more insight. Additionally, our study was limited to certain datasets and models like GPT-3.5 and GPT-4. Testing on more diverse tasks and newer models could reveal different trends. 


\bibliography{custom}

\begin{thebibliography}{24}
\providecommand{\natexlab}[1]{#1}

\bibitem[{Brown et~al.(2020)Brown, Mann, Ryder, Subbiah, Kaplan, Dhariwal, Neelakantan, Shyam, Sastry, Askell et~al.}]{brown2020language}
Tom Brown, Benjamin Mann, Nick Ryder, Melanie Subbiah, Jared~D Kaplan, Prafulla Dhariwal, Arvind Neelakantan, Pranav Shyam, Girish Sastry, Amanda Askell, et~al. 2020.
\newblock Language models are few-shot learners.
\newblock \emph{Advances in neural information processing systems}, 33:1877--1901.

\bibitem[{Cobbe et~al.(2021)Cobbe, Kosaraju, Bavarian, Chen, Jun, Kaiser, Plappert, Tworek, Hilton, Nakano et~al.}]{cobbe2021training}
Karl Cobbe, Vineet Kosaraju, Mohammad Bavarian, Mark Chen, Heewoo Jun, Lukasz Kaiser, Matthias Plappert, Jerry Tworek, Jacob Hilton, Reiichiro Nakano, et~al. 2021.
\newblock Training verifiers to solve math word problems.
\newblock \emph{arXiv preprint arXiv:2110.14168}.

\bibitem[{Fu et~al.(2023)Fu, Peng, Sabharwal, Clark, and Khot}]{fu2023complexitybased}
Yao Fu, Hao Peng, Ashish Sabharwal, Peter Clark, and Tushar Khot. 2023.
\newblock \href {https://arxiv.org/abs/2210.00720} {Complexity-based prompting for multi-step reasoning}.
\newblock \emph{Preprint}, arXiv:2210.00720.

\bibitem[{Geva et~al.(2021)Geva, Khashabi, Segal, Khot, Roth, and Berant}]{geva2021did}
Mor Geva, Daniel Khashabi, Elad Segal, Tushar Khot, Dan Roth, and Jonathan Berant. 2021.
\newblock Did aristotle use a laptop? a question answering benchmark with implicit reasoning strategies.
\newblock \emph{Transactions of the Association for Computational Linguistics}, 9:346--361.

\bibitem[{Jin et~al.(2024)Jin, Yu, Huang, Zeng, Wang, Hua, Zhao, Mei, Meng, Ding, Yang, Du, and Zhang}]{jin2024exploring}
Mingyu Jin, Qinkai Yu, Jingyuan Huang, Qingcheng Zeng, Zhenting Wang, Wenyue Hua, Haiyan Zhao, Kai Mei, Yanda Meng, Kaize Ding, Fan Yang, Mengnan Du, and Yongfeng Zhang. 2024.
\newblock \href {https://arxiv.org/abs/2404.07066} {Exploring concept depth: How large language models acquire knowledge at different layers?}
\newblock \emph{Preprint}, arXiv:2404.07066.

\bibitem[{Kojima et~al.(2023)Kojima, Gu, Reid, Matsuo, and Iwasawa}]{kojima2023large}
Takeshi Kojima, Shixiang~Shane Gu, Machel Reid, Yutaka Matsuo, and Yusuke Iwasawa. 2023.
\newblock \href {https://arxiv.org/abs/2205.11916} {Large language models are zero-shot reasoners}.
\newblock \emph{Preprint}, arXiv:2205.11916.

\bibitem[{Koncel-Kedziorski et~al.(2015)Koncel-Kedziorski, Hajishirzi, Sabharwal, Etzioni, and Ang}]{koncel2015parsing}
Rik Koncel-Kedziorski, Hannaneh Hajishirzi, Ashish Sabharwal, Oren Etzioni, and Siena~Dumas Ang. 2015.
\newblock Parsing algebraic word problems into equations.
\newblock \emph{Transactions of the Association for Computational Linguistics}, 3:585--597.

\bibitem[{Li et~al.(2023)Li, Sreenivasan, Giannou, Papailiopoulos, and Oymak}]{li2023dissecting}
Yingcong Li, Kartik Sreenivasan, Angeliki Giannou, Dimitris Papailiopoulos, and Samet Oymak. 2023.
\newblock Dissecting chain-of-thought: A study on compositional in-context learning of mlps.
\newblock \emph{arXiv preprint arXiv:2305.18869}.

\bibitem[{Ling et~al.(2017)Ling, Yogatama, Dyer, and Blunsom}]{ling2017program}
Wang Ling, Dani Yogatama, Chris Dyer, and Phil Blunsom. 2017.
\newblock Program induction by rationale generation: Learning to solve and explain algebraic word problems.
\newblock In \emph{Proceedings of the 55th Annual Meeting of the Association for Computational Linguistics (Volume 1: Long Papers)}, pages 158--167.

\bibitem[{Lyu et~al.(2023)Lyu, Havaldar, Stein, Zhang, Rao, Wong, Apidianaki, and Callison-Burch}]{lyu2023faithful}
Qing Lyu, Shreya Havaldar, Adam Stein, Li~Zhang, Delip Rao, Eric Wong, Marianna Apidianaki, and Chris Callison-Burch. 2023.
\newblock \href {https://arxiv.org/abs/2301.13379} {Faithful chain-of-thought reasoning}.
\newblock \emph{Preprint}, arXiv:2301.13379.

\bibitem[{Madaan and Yazdanbakhsh(2022)}]{madaan2022text}
Aman Madaan and Amir Yazdanbakhsh. 2022.
\newblock Text and patterns: For effective chain of thought, it takes two to tango.
\newblock \emph{arXiv preprint arXiv:2209.07686}.

\bibitem[{Merrill and Sabharwal(2023)}]{merrill2023expressive}
William Merrill and Ashish Sabharwal. 2023.
\newblock \href {https://arxiv.org/abs/2310.07923} {The expressive power of transformers with chain of thought}.
\newblock \emph{Preprint}, arXiv:2310.07923.

\bibitem[{OpenAI(2023)}]{openai2023gpt}
R~OpenAI. 2023.
\newblock Gpt-4 technical report. arxiv 2303.08774.
\newblock \emph{View in Article}, 2:13.

\bibitem[{Ouyang et~al.(2022)Ouyang, Wu, Jiang, Almeida, Wainwright, Mishkin, Zhang, Agarwal, Slama, Ray et~al.}]{ouyang2022training}
Long Ouyang, Jeffrey Wu, Xu~Jiang, Diogo Almeida, Carroll Wainwright, Pamela Mishkin, Chong Zhang, Sandhini Agarwal, Katarina Slama, Alex Ray, et~al. 2022.
\newblock Training language models to follow instructions with human feedback.
\newblock \emph{Advances in Neural Information Processing Systems}, 35:27730--27744.

\bibitem[{Patel et~al.(2021)Patel, Bhattamishra, and Goyal}]{patel2021nlp}
Arkil Patel, Satwik Bhattamishra, and Navin Goyal. 2021.
\newblock Are nlp models really able to solve simple math word problems?
\newblock In \emph{Proceedings of the 2021 Conference of the North American Chapter of the Association for Computational Linguistics: Human Language Technologies}, pages 2080--2094.

\bibitem[{Roy and Roth(2015)}]{roy2015solving}
Subhro Roy and Dan Roth. 2015.
\newblock Solving general arithmetic word problems.
\newblock In \emph{Proceedings of the 2015 Conference on Empirical Methods in Natural Language Processing}. Association for Computational Linguistics.

\bibitem[{Schaeffer et~al.(2023)Schaeffer, Miranda, and Koyejo}]{schaeffer2023emergent}
Rylan Schaeffer, Brando Miranda, and Sanmi Koyejo. 2023.
\newblock Are emergent abilities of large language models a mirage?
\newblock \emph{arXiv preprint arXiv:2304.15004}.

\bibitem[{Shao et~al.(2023)Shao, Gong, Shen, Huang, Duan, and Chen}]{shao2023synthetic}
Zhihong Shao, Yeyun Gong, Yelong Shen, Minlie Huang, Nan Duan, and Weizhu Chen. 2023.
\newblock \href {https://arxiv.org/abs/2302.00618} {Synthetic prompting: Generating chain-of-thought demonstrations for large language models}.
\newblock \emph{Preprint}, arXiv:2302.00618.

\bibitem[{Tang et~al.(2023)Tang, Zheng, Li, Meng, Zhu, Liang, and Zhang}]{tang2023large}
Xiaojuan Tang, Zilong Zheng, Jiaqi Li, Fanxu Meng, Song-Chun Zhu, Yitao Liang, and Muhan Zhang. 2023.
\newblock Large language models are in-context semantic reasoners rather than symbolic reasoners.
\newblock \emph{arXiv preprint arXiv:2305.14825}.

\bibitem[{Wang et~al.(2023)Wang, Wei, Schuurmans, Le, Chi, Narang, Chowdhery, and Zhou}]{wang2023selfconsistency}
Xuezhi Wang, Jason Wei, Dale Schuurmans, Quoc Le, Ed~Chi, Sharan Narang, Aakanksha Chowdhery, and Denny Zhou. 2023.
\newblock \href {https://arxiv.org/abs/2203.11171} {Self-consistency improves chain of thought reasoning in language models}.
\newblock \emph{Preprint}, arXiv:2203.11171.

\bibitem[{Wei et~al.(2022)Wei, Wang, Schuurmans, Bosma, Xia, Chi, Le, Zhou et~al.}]{wei2022chain}
Jason Wei, Xuezhi Wang, Dale Schuurmans, Maarten Bosma, Fei Xia, Ed~Chi, Quoc~V Le, Denny Zhou, et~al. 2022.
\newblock Chain-of-thought prompting elicits reasoning in large language models.
\newblock \emph{Advances in Neural Information Processing Systems}, 35:24824--24837.

\bibitem[{Wu et~al.(2023)Wu, Shen, Badrinath, Ma, and Lakkaraju}]{wu2023analyzing}
Skyler Wu, Eric~Meng Shen, Charumathi Badrinath, Jiaqi Ma, and Himabindu Lakkaraju. 2023.
\newblock Analyzing chain-of-thought prompting in large language models via gradient-based feature attributions.
\newblock \emph{arXiv preprint arXiv:2307.13339}.

\bibitem[{Yao et~al.(2023)Yao, Yu, Zhao, Shafran, Griffiths, Cao, and Narasimhan}]{yao2023tree}
Shunyu Yao, Dian Yu, Jeffrey Zhao, Izhak Shafran, Thomas~L. Griffiths, Yuan Cao, and Karthik Narasimhan. 2023.
\newblock \href {https://arxiv.org/abs/2305.10601} {Tree of thoughts: Deliberate problem solving with large language models}.
\newblock \emph{Preprint}, arXiv:2305.10601.

\bibitem[{Zhang et~al.(2022)Zhang, Zhang, Li, and Smola}]{zhang2022automatic}
Zhuosheng Zhang, Aston Zhang, Mu~Li, and Alex Smola. 2022.
\newblock Automatic chain of thought prompting in large language models.
\newblock \emph{arXiv preprint arXiv:2210.03493}.

\end{thebibliography}
\clearpage
\appendix

\onecolumn\section{Appendix}
\label{sec:appendix}
\begin{table}[!h]
\small
\caption{}
  \label{wide_table1}
  \begin{tabularx}{\textwidth}{X}
    \toprule
    \multicolumn{1}{c}{\textbf{Think About The Word}} \\\midrule
\textbf{prompt}: \\
Q: Could someone in Tokyo take a taxi to the The Metropolitan Museum of Art?\\
A: Let's think step by step. The stem of the sentence is Tokyo, take a taxi, Metropolitan Museum of Art. \textcolor{blue}{Think about Tokyo... Think about take a taxi... Think about Metropolitan Museum of Art...} Inference: Tokyo is in Japan and the Metropolitan Museum of Art is in New York. The two places are separated by the sea, so you can't take a taxi there. Since two places are separated by the sea, you can't take a taxi there. The answer is yes.  \\
      
Q: Could a chipmunk fit 100 chocolate chips in his mouth?\\\midrule
\textbf{Answer}:\\
A:
Based on this question, The stem of the sentence is chipmunk, fit, 100 chocolate chips, mouth. \textcolor{blue} {Think about chipmunk... Think about fit... Think about chocolate chips...} Inference: Chipmunks are small animals with relatively small mouths, so it's unlikely that a chipmunk would be able to fit 100 chocolate chips in its mouth at once. The answer is no.\\
\textbf{pred} : no\\
\textbf{GT} : no\\\toprule
    \multicolumn{1}{c}{\textbf{Read the question again}} \\\midrule
\textbf{prompt}: \\
Q: Mark's father gave him \$85. Mark bought 10 books, each of which cost \$5. How much money does Mark have left?\\
A: Let's think step by step. \textcolor{blue} {The question is: How much money does Mark have left? So we need to calculate How much money does Mark have left. Start looking for information about money now.} Mark's father gave him \$85. Mark bought 10 books, each of which cost \$5. That means that Mark spent \$50 on books. So we have equation money = +85 - 50 = 35. So Mark has \$85 - \$50 = \$35 left. So the answer is 35.\\
      
Q: During a visit to an orchard, Charlie picked 0.1666 of a bag of Golden Delicious apples, 0.1666 of a bag of Macintosh apples, and 0.333 of a bag of Cortland apples. How many bags of fruit did Charlie pick in total?\\
        \midrule
\textbf{Answer}:\\
A:
\textcolor{blue}{ The question is: How many bags of fruit did Charlie pick in total. So we need to calculate how many bags of fruit did Charlie pick in total. Start looking for information about bags now.} Charlie picked 0.166 of a bag of Golden Delicious apples, 0.166 of a bag of Macintosh apples, and 0.3333 of a bag of Cortland apples. We have to add these fractions together to find out how many bags of fruit Charlie picked in total. 0.1666 + 0.1666 + 0.3333 = 0.6666. This means that Charlie picked a total of 0.666 bags of fruit. The answer is 0.6666.\\
\textbf{pred} : 0.6666\\
\textbf{GT} : 0.6666\\
    \bottomrule

\end{tabularx}
\end{table}
\begin{table*}[htbp]
\small
  \caption{}
  \label{wide_table2}
  \begin{tabularx}{\textwidth}{X}
    \toprule
    \multicolumn{1}{c}{\textbf{Repeat State}} \\\midrule
\textbf{prompt}: \\
Q: A coin is heads up. Janette does not flip the coin. Stacey flips the coin. Ronny flips the coin. Kim does not flip the coin. Is the coin still heads up? Note that "flip" here means "reverse".\\
A: Let's think step by step. \textcolor{blue}{The state of the coin from the beginning is heads up.} Janette does not flip the coin, \textcolor{blue} {so the coin remains heads up. coin is heads up.} Stacey flips the coin, \textcolor{blue}{so the coin is now tails up. coin is now tail up.} Ronny flips the coin, so the coin is now heads up again. Kim does not flip the coin, \textcolor{blue} {so the coin remains heads up.} coin is head up. The answer is yes.
      
Q: A coin is heads up. Jeff does not flip the coin. Jen flips the coin. Giselle flips the coin. Noel does not flip the coin. Is the coin still heads up? Note that "flip" here means "reverse". \\
        \midrule
\textbf{Answer}:\\
\textcolor{blue}{The state of the coin from the beginning is heads up.} Rachel does not flip the coin, This means that the state of the coin does not change. \textcolor{blue}{so the coin remains heads up.} Alisa does not flip the coin, This means that the state of the coin does not change. \textcolor{blue}{so the coin remains heads up.} Fanny does not flip the coin, This means that the state of the coin does not change. so the coin remains heads up. Jeffery does not flip the coin, This means that the state of the coin does not change. \textcolor{blue}{so the coin remains heads up.} So the final answer is yes, the coin is still heads up. The answer is yes.\\
\textbf{pred} : yes\\
\textbf{GT} : yes\\

    \toprule
    \multicolumn{1}{c}{\textbf{Self-Verification}} \\
    \midrule
\textbf{prompt}: \\
Q: Take the last letters of each words in "Alina Alessandra Amina Bianca" and concatenate them.\\
A: Let's think step by step. \textcolor{blue} {There is four words.} So the answer will consist of four letters. Explain concatenate: concatenate is a term used in computer programming and mathematics, referring to the operation of combining two or more strings, sequences, or sets in a linear order. The last letter of "Alina" is "a". The last letter of "Alessandra" is "a". The last letter of "Amina" is "a". The last letter of "Bianca" is "a". \textcolor{blue} {So we have four letters.} So the final answer is "aaaa". The answer is aaaa.\\
Q: Take the last letters of each words in "Jeff Jen Giselle Noel" and concatenate them.\\
        \midrule
\textbf{Answer}:\\
 \textcolor{blue}{There is four words.} So the answer will consist of four letters. Explain concatenate: concatenate is a term used in computer programming and mathematics, referring to the operation of combining two or more strings, sequences, or sets in a linear order. The last letter of "Jeff" is "f". The last letter of "Jen" is "n". The last letter of "Giselle" is "e". The last letter of "Noel" is "l".  \textcolor{blue}{So we have four letters.} So the final answer is "fnel". The answer is fnel.\\
\textbf{pred} : fnel\\
\textbf{GT} : fnel\\

    \toprule
    \multicolumn{1}{c}{\textbf{Make Equation}} \\
    \midrule
\textbf{prompt}: \\
Q: 5 children were riding on the bus. At the bus stop 63 children got off the bus while some more got on the bus. Then there were 14 children altogether on the bus. How many more children got on the bus than those that got off?\\
A: Let's think step by step. first step, 5 children were riding on the bus. We know 5 children is on the bus. second step,There were 63 children that got off the bus. \textcolor{blue}{third step, some more got on the bus we define as unknown x.} fourth step, 14 children remained on the bus, which means we can calculate unknow x.\textcolor{blue}{we have equation x+5-63 = 14, now we know x is 72}. fifth step, Therefore, 72 - 63 = 9. 9 more children got on the bus than those that got off. The answer is 9.\\
Q: Rachel had to complete 8 pages of math homework. If she had to complete 3 more pages of math homework than reading homework How many pages did she have to complete in all?\\
\midrule
\textbf{Answer}:\\
Explain the problem. The problem is asking us to find out how many pages of homework Rachel had to complete in total given that she had to complete 8 pages of math homework and 3 more pages of math homework than reading homework. Find the necessary information. First, we know that Rachel had to complete 8 pages of math homework. \textcolor{blue}{Let's call this "M".} Second, we know that she had to complete 3 more pages of math homework than reading homework.\textcolor{blue}{ Let's call the number of reading homework pages "R".} Find the solution. Based on the information given, \textcolor{blue}{we can use the equation M = R + 3 to represent Rachel's math homework in terms of her reading homework.} So, if Rachel had to complete 8 pages of math homework, \textcolor{blue}{we can substitute M = 8 into the equation and solve for R. 8 = R + 3, R = 8 - 3, R = 5.} Therefore, Rachel had to complete 5 pages of reading homework. Now, we can find the total number of pages Rachel had to complete by adding the math and reading homework together.

Total = 8 (math) + 5 (reading)
Total = 13

So, in total, Rachel had to complete 13 pages of homework.\\
\textbf{pred} : 13\\
\textbf{GT} : 13\\\bottomrule
  \end{tabularx}
\end{table*}

\begin{table}[th]
\caption{}
\small
  \begin{tabularx}{\textwidth}{X}
  \label{tabel2}
    \\\toprule
    \multicolumn{1}{c}{\textbf{Example of Wrong Prompt}} \\\midrule
    Arithmetic Dataset\\\midrule

\textbf{prompt}: \\
Q: Joan has 10 books. Tom has 38 books.  How many books do they have\\
A: Let's think step by step. Joan has 10 books. Tom has 38 books. we have equation books = 10 +\textcolor{blue}{8} = 48. They have 10 + \textcolor{blue}{38} = 48 books together.\\\midrule
Commonsense Dataset\\\midrule
\textbf{prompt}: \\
Q: Could someone in Tokyo take a taxi to the The Metropolitan Museum of Art?\\
Let's think step by step. The stem of the sentence is Tokyo, take a taxi, Metropolitan Museum of Art. Explain Tokyo: Tokyo is the capital city of Japan and one of the most populous metropolitan areas in the world. Explain Metropolitan Museum of Art: is a art museums in New York City. \textcolor{blue}{Inference: Tokyo is in Japan and the Metropolitan Museum of Art is in New York. The two places are separated by the sea, so you can take a taxi there.}\\\midrule
Symbolic Dataset\\\midrule
\textbf{prompt}: \\
Q: Take the last letters of each words in  'Tim Candace Cecil Misael' and concatenate them.\\
A: Let's think step by step. Explain letters: letters can have various meanings depending on the context, such as Alphabetic Characters, Correspondence, Literature and Books. There is four words. So the answer will consist of four letters. The last letter of 'Tim' is 'm'. The last letter of 'Candace' is 'e'. The last letter of "Cecil" is 'l'. The last letter of "Misael" is "l". \textcolor{blue}{So we have four letters.} So the final answer would be \textcolor{blue}{"mel"}.\\\bottomrule
    \label{table2}

\end{tabularx}
\end{table}

\end{document}